\documentclass[sigconf,review=false]{acmart}

\AtBeginDocument{%
  \providecommand\BibTeX{{%
    \normalfont B\kern-0.5em{\scshape i\kern-0.25em b}\kern-0.8em\TeX}}}

\setcopyright{acmlicensed}
\copyrightyear{2018}
\acmYear{2024}
\acmDOI{XXXXXXX.XXXXXXX}

\acmConference[MM'24]{Make sure to enter the correct
  conference title from your rights confirmation email}{October 28 - November 1,
  2024}{Melbourne, Australia.}
\acmISBN{978-1-4503-XXXX-X/18/06}
\usepackage{bm}
\usepackage{multirow}
\usepackage{tabularx}
\usepackage{booktabs}

\begin{document}

\title{RoFIR: Distortion Vector Map Guided Transformer for Robust Fisheye Image Rectification}

\author{Zhaokang Liao}
\affiliation{%
  \institution{University of Science and Technology of China}
  \city{Hefei}
  \country{China}}
\email{lzk950803@mail.ustc.edu.cn}

\author{Hao Feng}
\affiliation{%
  \institution{University of Science and Technology of China}
  \city{Hefei}
  \country{China}}
\email{haof@mail.ustc.edu.cn}

\author{Shaokai Liu}
\affiliation{%
  \institution{University of Science and Technology of China}
  \city{Hefei}
  \country{China}}
\email{liushaokai@mail.ustc.edu.cn}

\author{Wengang Zhou}
\affiliation{%
  \institution{University of Science and Technology of China}
  \city{Hefei}
  \country{China}}
\email{zhwg@mail.ustc.edu.cn}

\author{Houqiang Li}
\affiliation{%
  \institution{University of Science and Technology of China}
  \city{Hefei}
  \country{China}}
\email{lihq@mail.ustc.edu.cn}

\renewcommand{\shortauthors}{author name and author name, et al.}

\begin{abstract}
Fisheye images are categorized into central fisheye images and deviated fisheye images based on the optical center position. Existing rectification methods are limited to central fisheye images, while this paper proposes a novel method that extends to deviated fisheye image rectification. The challenge lies in the variant global distortion distribution pattern caused by the random optical center position. To address this challenge, we propose a distortion vector map (DVM) that measures the degree and direction of local distortion. By learning the DVM, the model can independently identify local distortions at each pixel without relying on global distortion patterns. The model adopts a pre-training and fine-tuning training paradigm. In the pre-training stage, it predicts the distortion vector map and perceives the local distortion features of each pixel. In the fine-tuning stage, it predicts a pixel-wise flow map for deviated fisheye image rectification. We also propose a data augmentation method mixing central, deviated, and distorted-free images. Such data augmentation promotes the model performance in rectifying both central and deviated fisheye images, compared with models trained on single-type fisheye images. Extensive experiments demonstrate the effectiveness and superiority of the proposed method.

\end{abstract}





\maketitle

\section{Introduction}
\par The fisheye camera refers to an ultra-wide-angle camera with a field of view~(FoV) close to 180 degrees. The expansive FoV inherent in fisheye cameras renders wide utilization in surveillance~\cite{muhammad-et-al:efficient,eichenseer-et-al:Motion}, autonomous driving~\cite{abu2018augmented,geiger2012we,grigorescu2020survey}, and virtual reality (VR)~\cite{wei2022towards,LinPolycamera2019}. However, ultra-wide FoV also inevitably leads to significant distortions in fisheye images. Such distortion diminishes the performance of mainstream computer vision tasks such as object detection~\cite{hsieh2023object,zhao2023distortion} and scene segmentation~\cite{liu2023caris,wang2016objectness}. As a result, a multitude of fisheye image rectification methods have emerged.

\begin{figure}[!t]
    \centerline{\includegraphics[width=1\columnwidth]{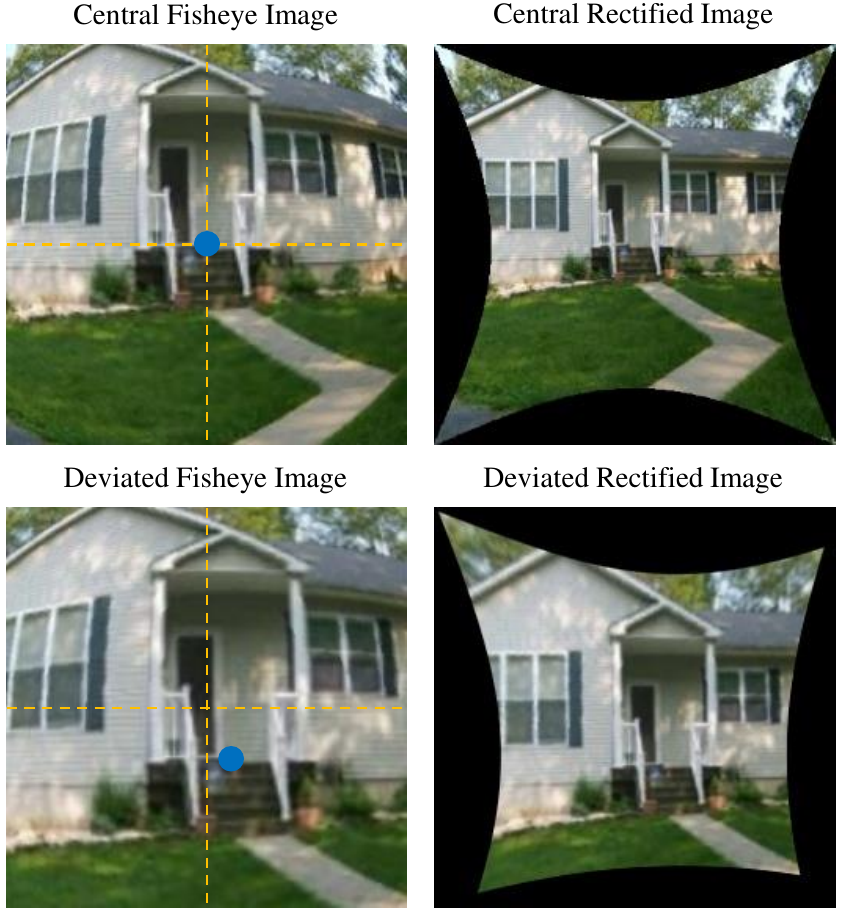}}
    \caption{Examples of a central fisheye image and a deviated fisheye image, together with their rectified images.
}
    \label{fig1}
\end{figure}

\par Fisheye image rectification methods can be generally categorized into two types, traditional methods and learning-based methods. Traditional methods rectify fisheye images according to human-derived knowledge in the image processing field. However, these methods often face challenges in generalizing to other fisheye cameras with varying physical parameters and sometimes depend on calibrated devices. Learning-based methods allow models to extract and identify fisheye distortion features, thus achieving distortion removal without human intervention. In recent years, the community has been impressed by the learning-based methods due to their minimal application constraints, excellent performance, and robust generalization ability. In detail, the learning-based methods can be further divided into three branches, including parameter-based methods, generation methods, and flow-based methods. Parameter-based methods leverage neural networks to predict distortion parameters of a fisheye image based on the division~\cite{fitzgibbon2001simultaneous} or polynomial~\cite{kannala-brandt:generic} distortion model. Generation methods adopt an encoder-decoder architecture to directly reconstruct the rectified image. Flow-based
method utilizes neural networks to predict the pixel-wise coordinate mapping between the fisheye image and the rectified image.

\par Existing rectification methods are limited to the fisheye images exhibiting radial symmetry. Nevertheless, according to our observations on real surveillance videos and images, the radially symmetric distortion pattern does not consistently hold, referencing the geometric center. The causation of this phenomenon is multifaceted. Firstly, the optical center, which is the intersection of the optical axis and the image plane, may not align with the geometric center of the image. Secondly, fisheye images may be incompletely displayed or asymmetrically cropped, a common occurrence in image post-processing. As a result, we present a new problem, how can fisheye images with deviated optical centers be rectified?

\par Formally, we categorize fisheye images into central fisheye images and deviated fisheye images based on the position of the optical center. In central fisheye images, the optical center is anchored at the geometric center of the image, whereas in deviated fisheye images, the location of the optical center is arbitrary. For the deviated fisheye image and the central fisheye image, we provide an example of each, as shown in Figure~\ref{fig1}. Rectifying deviated fisheye images is challenging for existing fisheye rectification methods. Parameter-based methods struggle to determine the rectified image from distortion parameters, owing to the unknown position of the optical center. Generation methods exhibit limitations in predicting the boundaries of rectified images and lower quality generation near the boundaries. This challenge is compounded when generating the non-symmetric boundaries in the rectified result of a deviated fisheye image. Flow-based methods rely on global distortion patterns in central fisheye images to promote the performance of these models, which are no longer valid in deviated fisheye images.

\par To address these challenges, we shift our approach from learning global distortion distributions to perceiving local distortions. Consequently, we propose the Distortion Vector Map(hereinafter referred to as $\bm{V}$ map or $\bm{V}$ label), which intuitively and quantitatively measures the local distortion magnitude and direction of a fisheye image at the pixel level. Each pixel of a fisheye image is represented by a two-dimensional vector in the $\bm{V}$ map. The magnitude of the vector indicates the degree of distortion at that point. The direction of the vector extends from the optical center of the fisheye image toward the current pixel. The $\bm{V}$ map is independent from specific fisheye camera models and distortion parameters. Additionally, because it includes distortion direction, the $\bm{V}$ map remains unaffected by the optical center deviation, making it suitable for evaluating both central and deviated fisheye images.

\par To implement our approach, we adopt a Vision Transformer(ViT) model~\cite{dosovitskiy2020image} and a pre-training and fine-tuning learning paradigm~\cite{he2022masked,devlin2018bert}. In the pre-training stage, our model takes fisheye images as input and predicts the $\bm{V}$ map, aiming at awareness of the local distortion patterns. After pre-training, we preserved a subset of the model weights for the fine-tuning stage. In the fine-tuning stage, our model predicts the pixel-wise flow map, given the input fisheye image. Ultimately, we construct the rectified image by applying bilinear sampling to the input fisheye image based on the pixel-wise flow map. Additionally, we propose a data augmentation method that integrates central and deviated fisheye images with distortion-free images in the dataset, thereby enhancing the model performance in rectifying both central and deviated fisheye images.

\par To evaluate the validity of our approach, we extensively evaluate our model on central and deviated fisheye image rectification, confirming its state-of-the-art performance in both scenarios, compared to the existing rectification methods.
\par In summary, this research presents three primary contributions as follows, 

\begin{itemize}
    \item 
    We identify a previously overlooked issue: the rectification of fisheye images with a deviated optical center.
    \item 
    We introduce the $\bm{V}$ map, which measures both the magnitude and direction of local distortion. The $\bm{V}$ map is suitable for measuring both central and deviated fisheye images.
    \item
    Extensive experiments demonstrate that our method can rectify deviated fisheye images and achieves state-of-the-art performance on both deviated and central fisheye images. 
\end{itemize} 

\section{Related Work}
\par Recently, the rectification of fisheye images has garnered significant interest within the academic community. This section offers an overview of fisheye image rectification methods, classified into two categories: traditional methods and learning-based methods.

\subsection{Traditional Methods}

\par Traditional methods rectify fisheye images according to human-derived knowledge. Traditional methods of one category \cite{barreto-daniilidis:fundamental,kukelova-pajdla:minimal,chander-et-al:summary,gasparini2009plane} rectifies fisheye distortion by identifying corresponding feature points from multiple perspectives. However, these methods rely on dedicated scenarios and hardware devices.
\par Other traditional methods \cite{bukhari-dailey:automatic,thormahlen-et-al:robust,wang-et-al:simple,geiger2012automatic} recognize and rectify distortion by detecting straight lines, utilizing the principle that straight lines bend into curves in fisheye images. Nevertheless, some fisheye images lack straight lines for effective rectification, such as natural landscapes and portraits. Additionally, minor errors in line detection lead to significant inaccuracies in the rectified image. Therefore, recent research trends favor robust, accurate, and hardware-free learning-based methods over traditional ones.

\subsection{Learning-based Methods}

Learning-based methods refer to the application of neural networks~\cite{he2016deep} for fisheye image rectification. Learning-based methods predominantly fall into three branches: parameter-based methods, generation-based methods, and flow-based methods.
\par The parameter-based methods utilize neural networks to deduce distortion parameters of fisheye images, which are then applied via a fixed function to calculate the rectified image. Rong~\cite{rong-et-al:radial} pioneered the training of a convolutional neural network for predicting distortion parameters. FisheyeRecNet~\cite{yin-et-al:fisheyerecnet} introduced a multi-content collaborative network, that estimates distortion parameters by processing high-level and low-level features independently. CSL~\cite{xue-et-al:learning} enhanced the efficiency and accuracy of the model in distortion parameter prediction by integrating prior geometric constraints.
\par The generation-based methods trained their models to directly predict the rectified images. DR-GAN \cite{liao-et-al:dr}, a variant of Generative Adversarial Network (GAN)~\cite{ledig-et-al:photo}, is employed for generating rectified images. PCN~\cite{yang-et-al:progressively} adopted an encoder-decoder architecture to progressively rectify the distortions in the fisheye image at multiple feature scales. ModelFree~\cite{liao-et-al:model} introduced a rectification framework anchored by a distortion distribution map independent of specific distortion models, which intuitively represents the pixel-level distortion magnitude, independent of specific distortion models. Other generation-based methods include Multi~\cite{liao-et-al:multi}, Polar~\cite{zhao-et-al:Polar} and Dynamic~\cite{liao-et-al:dynamic}, which are not further elaborated here.
\par The flow-based methods predict a pixel-wise mapping flow~\cite{feng2021doctr,xu2022gmflow} between a fisheye image and a rectified one. These models, including DaFIR~\cite{liao-et-al:DaFIR}, SimFIR~\cite{feng-et-al:simfir}, and RDTR~\cite{wang-et-al:Model-aware} can be improved by a pre-training stage with different pretext tasks. 

\section{Preliminaries}

This section firstly presents the camera model of fisheye distortion and then introduces how a proposed distortion vector map measures the local fisheye distortion.
\subsection{Camera Model of Fisheye Distortion}\label{camera}
Image formation involves projecting 3D spatial coordinates onto a 2D plane via a camera model. The lens assembly of a fisheye camera is regarded as a nonlinear optical system, commonly approximated by a polynomial relationship. According to the fisheye camera model, the polynomial relationship \cite{kannala-brandt:generic} between the incidence angle $\theta_c$ and the emergence angle $\theta_d$ is formulated as follows,
\begin{equation}
    \theta_c = \sum_{i=1}^\infty {\lambda_i}{\theta_d^{2i-1}}      ,            
    \label{eq1}
\end{equation}
where $\theta_c$ symbolizes the incident light angle, while $\theta_d$ signifies the emergent light angle. $\lambda_i$ is the distortion parameter, used to characterize the fisheye lens distortion.
\par Utilizing the equidistant projection model in pinhole cameras leads to the formulations $r_c = f \cdot tan\theta_c$ and $r_d = f \cdot tan\theta_d$. Here, $f$ represents the focal length of the camera. The term $r_c$ is the distance from a point $P$ in the rectified image to its optical center. Similarly, $r_d$ indicates the distance from the optical center of the distorted image to the point $P'$, which corresponds to $P$. 
In a central fisheye image, the optical center coincides with the geometric center. However, in a deviated fisheye image, there exists a deviation between the geometric center and the optical center.  For simplicity, $r_c$ and $r_d$ can be approximated as $r_c = f\cdot tan\theta_c\approx f\cdot\theta_c$ and similarly $r_d = f\cdot tan\theta_d\approx f\cdot\theta_d$. By multiplying both sides of Eq.~\eqref{eq1} by $f$, the equation is transformed to,
\begin{equation}
    r_c=f\sum_{i=1}^\infty {\lambda_i}{\theta_d^{2i-1}} 
       .
    \label{eq2}
\end{equation}
Parameter $f$ remains constant and is not influenced by $\theta_d$. By defining $k_i=\frac{\lambda_i}{f^{2i-2}}$, Eq.~\eqref{eq2} can be written as
\begin{equation}
    r_c=\sum_{i=1}^\infty k_if^{2i-1}{\theta_d^{2i-1}} 
       .
    \label{eq3}
\end{equation}
By substituting the approximation $r_d = f\cdot tan\theta_d\approx f\cdot\theta_d$ into the Eq.~\eqref{eq3} and integrating, we derive the polynomial model for fisheye distortion as follows,
\begin{equation}
    r_c=\sum_{i=1}^\infty k_ir_d^{2i-1} 
       .
    \label{distortion_model}
\end{equation}
\begin{figure}[t]
	\centering
	\includegraphics[width=0.9\columnwidth]{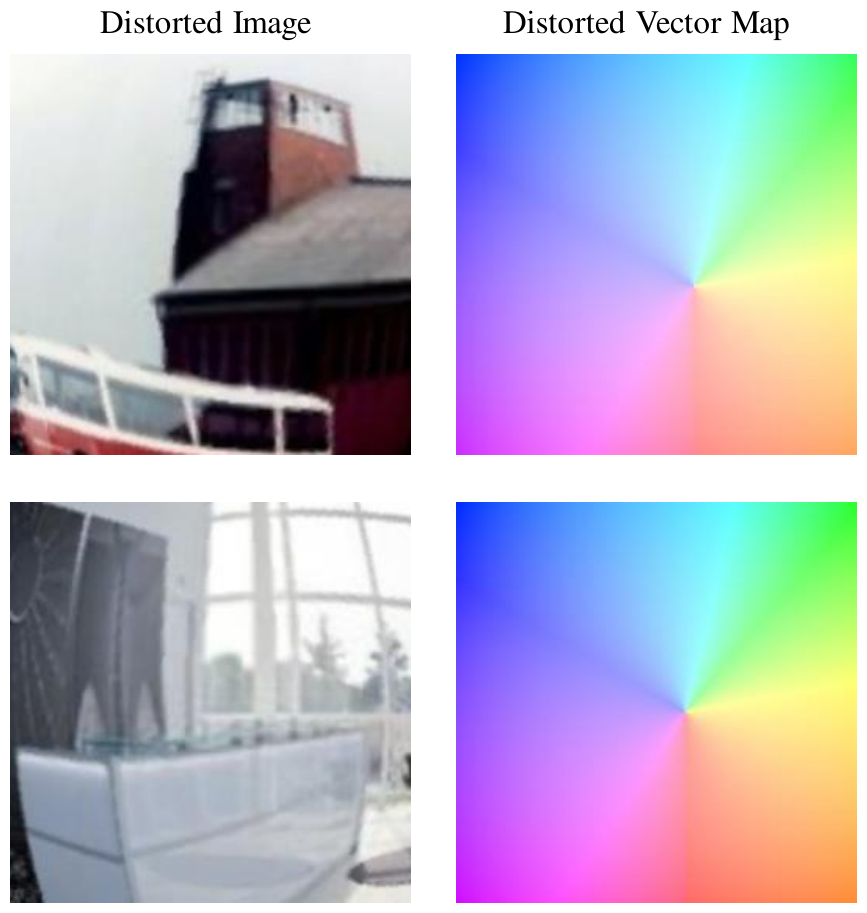}
	\caption{Distortion patterns of two fisheye images are illustrated by their Distortion Vector Maps.}
	\label{fig:ddm}
 	\vspace{-0.1in}
\end{figure}
\subsection{Distortion Vector Map}
Motivated by the shortcomings of existing methods in rectifying deviated fisheye images, we propose the $\bm{V}$ map that quantifies the local distortion magnitude and direction for both the deviated and the central fisheye images at the pixel level. Formally, for any given fisheye image, each pixel value in its $\bm{V}$ map is defined as follows, 
 \begin{equation}
     {\texttt{$\bm{V}$}}(x_d,y_d) = \frac{r_d(x_d-x_{od},  y_d-y_{od})}{r_c\sqrt{(x_d-x_{od})^{2}+(y_d-y_{od})^{2}}} , 
     \label{dvm}     
\end{equation}
where $(x_d,y_d)$ represents the coordinate of a point in the distorted image, and $(x_{od},y_{od})$ denotes the coordinate of the optical center point in the distorted image. According to Section~\ref{camera}, we have, 
 \begin{equation}
   \frac{r_d}{r_c} = \frac{\sqrt{(x_d-x_{od})^{2}+(y_d-y_{od})^{2}}}{\sqrt{(x_c-x_{oc})^{2}+(y_c-y_{oc})^{2}}}, 
   \label{dvm2}
\end{equation}
where $(x_c,y_c)$ in the rectified image represents the corresponding point of $(x_d,y_d)$ in the distorted image. Similarly, $(x_{oc},y_{oc})$ represents the corresponding point of $(x_{od},y_{od})$. Eq.~\eqref{dvm} and Eq.~\eqref{dvm2} indicate that each pixel in the $\bm{V}$ corresponds to a two-dimensional vector. The magnitude of this vector equals $r_d/r_c$, representing the distortion degree. The direction of the vector points from the current pixel toward the optical center of the distorted image. Specifically, Figure~\ref{fig:ddm} presents the $\bm{V}$ maps of two fisheye image examples. 
 
\par The $\bm{V}$ map quantifies the magnitude and direction of local distortions in both central and deviated fisheye images. That the $\bm{V}$ remains robust to the deviation of the optical center stems from the fact that both the distortion direction and the distance related to distortion magnitude are relative metrics, using the optical center as a reference. Values of the $\bm{V}$ map remain independent of the absolute position in the distorted image. Furthermore, the proposed $\bm{V}$ applies to any fisheye camera model. Moreover, the V map can be efficiently produced. For a given fisheye distortion image, the time required to annotate the V label is merely one-sixtieth of that needed for annotating the pixel-wise flow map.

\begin{figure}[t]
	\centering
	\includegraphics[width=1\columnwidth]{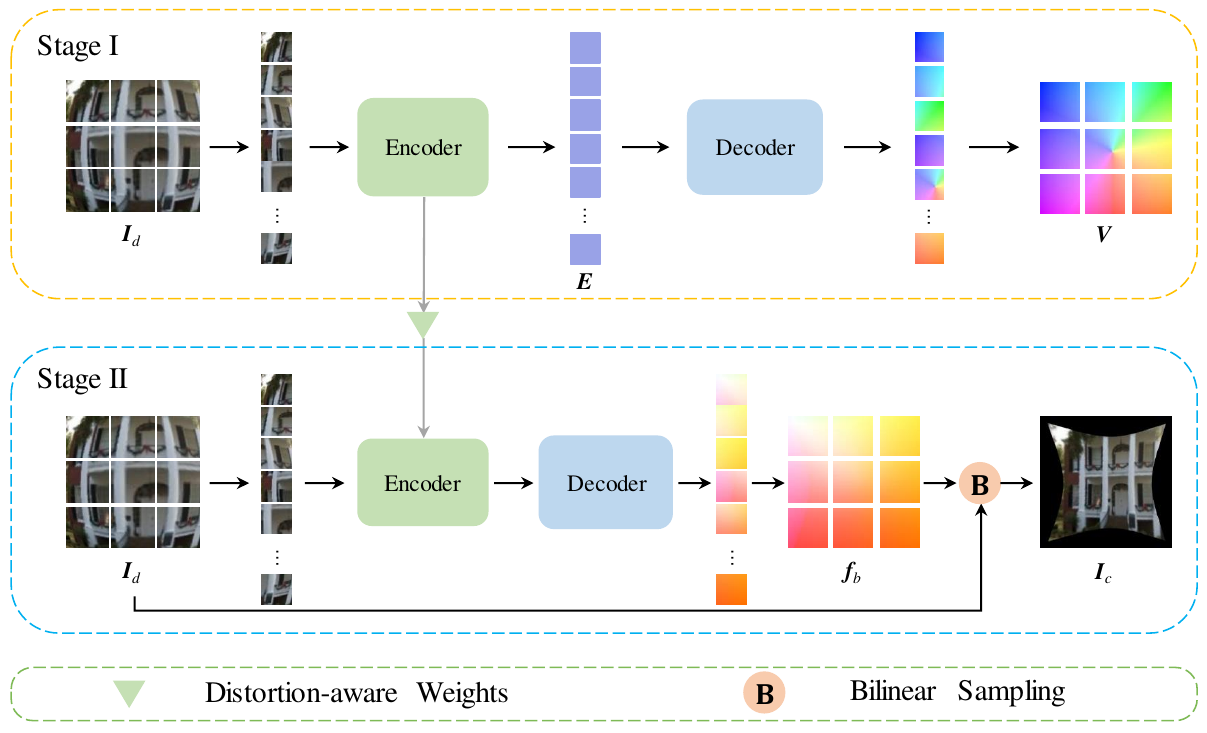}
	\caption{Framework of our method for fisheye image rectification. 
	It consists of two stages: 
	(I) supervised pre-training that learns the local distortion features of fisheye images through a DVM estimation pretext task.
	(II) fine-tuning for rectification which leverages the learned representation to reconstruct the rectified image with a pixel-wise flow map.
	}
	\label{framework}
	\vspace{-0.1in}
\end{figure}

\section{Method}

This section introduces a robust rectification framework~(RoFIR) for fisheye images, which recognizes the local distortion features and predicts the pixel-wise flow map for rectification. Figure~\ref{framework} illustrates the RoFIR model architecture, comprising a RoFIR encoder and decoder. The training of the RoFIR model consists of two stages: (a) Pre-training to regress the $\bm{V}$ label for local distortion perception, and (b) Fine-tuning for accurate prediction of the pixel-wise flow map. These stages are detailed in the following sections.

\subsection{Pre-training for Distortion Learning}
This section introduces the pre-training stage for the RoFIR model. Due to the lack of inductive bias, we design a pre-training stage adopting $\bm{V}$ label regression as a pretext task, enhancing the representation of distortion in central and deviated fisheye images. 
\par The process of the pre-training stage consists of three steps. First, the input fisheye image is segmented into non-overlapping square patches of equal size. Subsequently, in a process known as tokenization, the RoFIR encoder linearly projects these image patches into embedding vectors and adds position embeddings. The token embeddings derived from the input fisheye image patches pass through the RoFIR encoder and decoder, and the model regresses the distortion vector map. Finally, the RoFIR encoder weights are preserved, whereas the RoFIR decoder weights are relinquished.

\smallskip
\par
\textbf{Segmentation of the Input Image.}
 The standard transformer encoder architecture \cite{vaswani-et-al:attention} necessitates token embeddings as input. Consequently, the initial step involves segmentation of the fisheye image $\bm{I}_d \in\mathbb{R}^{H\times W\times 3}$ into a sequence of patches $\bm{x}_p \in\mathbb{R}^{N\times (P^2\times 3)}$ by row order, where $H$ and $W$ denote the height and width of the image $\bm{I}_d$ respectively. $P$ represents the side length of the square patches while $N=HW/P^2$ denotes the total count of patches. 

\smallskip
\par
\begin{figure}[t]
	\centering
	\includegraphics[width=1\columnwidth]{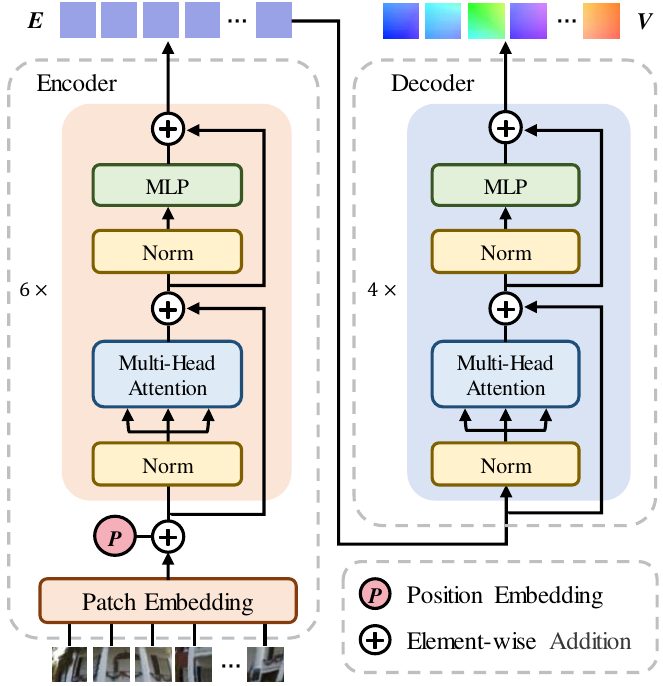}
	\caption{The internal structure of the DaFIR encoder and the DaFIR decoder. }
	\label{architecture}
	\vspace{-0.1in}
\end{figure}

\textbf{RoFIR Encoder.}
Figure~\ref{architecture} illustrates that the RoFIR encoder consists of a patch embedding layer, a position embedding layer, and $6$ transformer encoder \cite{vaswani-et-al:attention} layers. First, the image patches $\bm{x}_p$ are linearly projected into embedding vectors $\bm{E} \in\mathbb{R}^{N\times D}$. Subsequently, absolute position embedding vectors of the same dimensions are added to the embedding vectors $\bm{E}$, to preserve positional information of each patch. This step transfers the fisheye image patches into token embeddings $\bm{T} \in\mathbb{R}^{N\times D}$. Finally, the token embeddings $\bm{T}$ pass through $6$ transformer encoder layers within the RoFIR encoder. The RoFIR encoder is engineered to capture the local distortion features of both deviated and central fisheye images.

\smallskip
\par
\textbf{RoFIR Decoder.}
Following the RoFIR encoder, as depicted in Figure~\ref{architecture}, the RoFIR decoder incorporates $4$ transformer encoder layers. The RoFIR decoder is responsible for the prediction of the $\bm{V}$ according to the distortion features. The feature passing through the terminal layer of the RoFIR decoder is linearly projected to $\mathbb{R}^{N \times \frac{2D}{3}}$, as the dimension of $V \in\mathbb{R}^{H \times W \times 2}$. Finally, the feature is resized from the dimension $\mathbb{R}^{N \times \frac{2D}{3}}$ to $\mathbb{R}^{H\times W\times 2}$, which matches the dimension of $\bm{V}$ for output and can be regarded as the reverse process of patch segmentation. 

\smallskip
\par
\textbf{Preservation of Partial Weights.} 
After pre-training, the weights of the RoFIR encoder are preserved, whereas the weights of the RoFIR decoder are relinquished. The preserved weights are employed to initialize the RoFIR encoder in the fine-tuning stage.

\smallskip
\par
\textbf{Training Loss.} 
The pre-training stage is optimized end-to-end, governed by the following training loss:
\begin{equation}
	\mathcal{L}_{d} = \left \| \bm{V}_{pre} - \bm{V}_{gt} \right \|_2,
\end{equation}
where $\mathcal{L}_{d}$ is the  $L_2$ distance between the predicted $\bm{V}_{pre}$ and given ground truth $\bm{V}_{gt}$.
\subsection{Fine-tuning for Distortion Rectification}
This section details the second stage of RoFIR, \emph{i.e.,} fine-tuning for rectification. Initially, the RoFIR encoder is initialized by the weights derived from the pre-trained RoFIR encoder, and the RoFIR decoder is initialed randomly. Subsequently, the model processes the input fisheye image to predict a pixel-wise flow map. Eventually, utilizing a fixed bilinear sampling function, this flow map guides the sampling of the input image to construct the rectified image.

\smallskip
\par
\textbf{Distortion-aware Weights Initialization.}
After pre-training, the RoFIR encoder can perceive local distortion features and is robust against optical center shifts. As a result, in the fine-tuning stage, the RoFIR encoder is initialized with the weights preserved from its pre-training stage, whereas the RoFIR decoder is initialized randomly. This strategy results in a model proficient in rectifying both central and deviated fisheye images.

\smallskip
\par
\textbf{Prediction of Pixel-wise Flow Map.}
This paragraph introduces the pixel-wise flow map and discusses its advantages. The pixel-wise flow map delineates the correspondence between pixels of the input and output images. Through the pixel-wise flow map, distortion rectification is formulated as a sampling of the distorted image. In this case, the model does not need to reconstruct image content, thereby maximizing the preservation of detail in distorted images. Specifically, such maps are divided into two categories: forward and backward flow maps. Blind~\cite{li-et-al:blind} employs a forward flow map for unwrapping distorted images. However, this method encounters issues where some target pixels lack corresponding source pixels, resulting in cracks in the rectified image. Consequently, Blind utilizes the Hough Transform technique to address the issue of cracks. In the backward flow map, each target pixel corresponds to a source pixel. As a result, our approach employs a backward flow map to inherently avoid the cracks issue. The specific laws of the backward flow map are elucidated next.
\par Given a fisheye image as input, the DaFIR model is trained to predict a pixel-wise flow map $\bm{f}_b \in\mathbb{R}^{H \times W \times 2}$. The pixel-wise flow map indicates the corresponding target pixel in the rectified image $\bm{I}_c \in\mathbb{R}^{H \times W \times 3}$ for each source pixel. The pixel $\bm{I}_c(u,v)$ in the rectified image corresponds to the pixel $ \bm{I}_d(\bm{f}_x(u,v), \bm{f}_y(u,v))$ in the distorted image, as shown in the formula:  
\begin{equation}\label{equ:task}
	\bm{I}_c(u,v) = \bm{I}_d(\bm{f}_x(u,v), \bm{f}_y(u,v)).
\end{equation}

\smallskip
\par
\textbf{Construction of the Rectification Image.}
The rectified image is generated by applying the bilinear sampling function to the input fisheye image with the pixel-wise flow map. According to the sampling laws, the pixel value $\bm{I}_c(u,v)$ in the rectified image equals $\bm{I}_d(\bm{f}_x(u,v), \bm{f}_y(u,v))$, if the coordinate $(\bm{f}_x(u,v), \bm{f}_y(u,v))$ is in the range of $\mathbb{R}^{2}: [0,H-1] \times [0, W-1]$. Otherwise, the pixel value $\bm{I}_c(u,v)$ in the rectified image is set to $(0,0,0)$.

\smallskip
\par
\textbf{Training Loss.} 
The fine-tuning stage is also optimized end-to-end, governed by the following training loss:
\begin{equation}
	\mathcal{L}_{f} = \left \| \bm{f}_{bpre} - \bm{f}_{bgt} \right \|_1,
\end{equation}
where $\mathcal{L}_{f}$ denotes the $L_1$ distance between the predicted pixel-wise flow map $\bm{f}_{bpre}$ and its ground truth $\bm{f}_{bgt}$.

\section{Experiment}
\par This section delineates the implementation setup, ablation studies, and comparison with other methods. Experimental results substantiate the efficacy of the RoFIR method and its superior performance in the rectification of both central and deviated fisheye images.

\subsection{Implementation Setup}\label{makedeviated}
This section outlines the implementation setup of our method, encompassing the construction of the deviated fisheye images, evaluation metrics, and hyperparameters.

\begin{table*}[t] 
\centering
\caption{Ablation studies about domain adaptation. C denotes the dataset in which all fisheye images are central fisheye images. D denotes the dataset in which all fisheye images are transferred to the deviated fisheye images. $\uparrow$ indicates the higher the better, while $\downarrow$ indicates the lower the better.}
\label{table:aba_domain}
\begin{tabularx}{4.8in}{cc|>{\centering\arraybackslash}X>{\centering\arraybackslash}X|>{\centering\arraybackslash}X>{\centering\arraybackslash}X}
\toprule
\multicolumn{2}{c|}{\textbf{Dataset}} & \multicolumn{2}{c|}{\textbf{Central Fisheye Image}} & \multicolumn{2}{c}{\textbf{Deviated Fisheye Image}} \\
\cmidrule{1-6}
\textbf{Pre-training} & \textbf{Fine-tuning} & \textbf{SSIM↑} & \textbf{PSNR↑} & \textbf{SSIM↑} & \textbf{PSNR↑} \\
\midrule
120k-C & 30k-C & \textbf{0.9053} & \textbf{26.05} & 0.5835 & 15.72 \\
280k-D & 70k-D & 0.7469 & 18.42 & \textbf{0.8905} & \textbf{23.45} \\
\bottomrule
\end{tabularx}
\end{table*}

\begin{table*}[!htbp] 
\centering
\caption{The ablation study about the pre-training stage. M denotes the dataset contains central fisheye images in which each batch has a 70\% chance to be transferred to deviated fisheye images.}
\label{table:aba_dvm}
\begin{tabularx}{4.8in}{cc|>{\centering\arraybackslash}X>{\centering\arraybackslash}X|>{\centering\arraybackslash}X>{\centering\arraybackslash}X}
\toprule
\multicolumn{2}{c|}{\textbf{Dataset}} & \multicolumn{2}{c|}{\textbf{Central Fisheye Image}} & \multicolumn{2}{c}{\textbf{Deviated Fisheye Image}} \\
\cmidrule{1-6}
\textbf{Pre-training} & \textbf{Fine-tuning} & \textbf{SSIM↑} & \textbf{PSNR↑} & \textbf{SSIM↑} & \textbf{PSNR↑} \\
\midrule
- & 100k-M & 0.8785 & 23.80 & 0.8343 & 21.70 \\
400k-M & 100k-M & \textbf{0.9097} & \textbf{25.92} & \textbf{0.8944} & \textbf{23.89} \\
\bottomrule
\end{tabularx}
\end{table*}

\vspace{0.1in}
\smallskip
\par
\textbf{Dataset.}
Given the absence of real fisheye image datasets and the complexity involved in calibrating real fisheye images to obtain ground truths, we adopt the approach delineated in SimFIR~\cite{feng-et-al:simfir} for dataset generation. To enhance the ability to identify distortion, we mix 20\% distortion-free images in all the datasets with fisheye images. The deviated fisheye images are transferred from central fisheye images dynamically during training. 
\par Specifically, the central fisheye images are generated from the distortion-free images in the Place365 dataset~\cite{zhou-et-al:places}. We utilize the polynomial model~\cite{kannala-brandt:generic} with the first four distortion parameters ($k1, k2, k3, k4$), adequately fitting most real-world application scenarios. Our method necessitates three datasets: a pre-training dataset, a fine-tuning dataset, and a testing dataset. The pre-training dataset contains 400K central fisheye images and 80K distortion-free images, together with the corresponding $\bm{V}$ labels. The values in the $\bm{V}$ label of a distortion-free image are unit vectors pointing towards the geometric center. For fine-tuning, an additional dataset of 100K distorted fisheye images and 20K distortion-free images, alongside their pixel-wise flow maps, is constructed. The pixel-wise flow map value of a pixel in the distortion-free image equals its coordinates. For the test dataset, we generate one dataset containing 5K central fisheye images and the other one containing 5K deviated fisheye images to evaluate the performance of our model.
\par A deviated fisheye image is constructed by cutting a square of random size in a central fisheye image from a random position. The $\bm{V}$ label and pixel-wise flow map of the deviated fisheye image are generated in the same way. The deviated fisheye image and its labels are generated dynamically during pre-training and fine-tuning, in which each epoch of central data has a 70\% chance of being generated to deviated data before being fed into the model.

\smallskip
\par
\textbf{Evaluation Metrics.}
For quantitative comparison between existing methods and RoFIR, we employ established metrics: Structural Similarity (SSIM) and Peak Signal to Noise Ratio (PSNR). Notably, SSIM is used to gauge the precision of structure rectification and PSNR evaluates the detail quality of the rectified images.

\smallskip
\par
\textbf{Hyperparameters.}
The spatial dimension of the image $(H, W)$ is set to $(256, 256)$, applicable to both the pixel-wise flow map and $D$ labels. The embedding dimension $D$ is established at 512. We use the Adam optimizer\cite{kingma2014adam} with a learning rate up to $10^{-4}$\cite{teed-Deng:raft}, training on two NVIDIA GeForce GTX 3090 GPUs. 

\subsection{Ablation Studies}
To ascertain the efficacy of the pre-training task involving $\bm{V}$ label prediction and the inclusion of distortion-free images in the datasets, we undertake a series of ablation studies.

\smallskip
\par
\textbf{Domain Adaptation Ability.}
To substantiate the necessity of employing a hybrid dataset comprising both central and deviated fisheye images, we investigate the domain adaptation capabilities of the RoFIR. Specifically, the RoFIR model trained on the central fisheye images is adopted to rectify the deviated fisheye images, and vice versa. The empirical outcomes, summarized in Table~\ref{table:aba_domain}, reveal no significant domain adaptation ability. This underscores the inability of models trained exclusively on central fisheye images to generalize to the rectification of distorted fisheye images.

\smallskip
\par
\textbf{Pre-training Stage.}
The efficacy of the pre-training strategy, regressing the $\bm{V}$ label, is appraised through ablation studies, detailed in Table~\ref{table:aba_dvm}. Absent pre-training, the baseline model exhibits acceptable performance. The model attains a PSNR of 24.36 and an SSIM of 0.8847 for central fisheye images, alongside a PSNR of 21.86 and an SSIM of 0.8338 for deviated fisheye images, attributable to the advantages of the pixel-wise flow map. By regressing the $\bm{V}$ label, the model identifies the local distortion much better, hence achieving significantly higher and more robust performances, especially for deviated fisheye image rectification.

\smallskip
\par
\textbf{Inclusion of the Distortion-free Image.}
To fortify the model's resilience against variance and improve its robustness, we intersperse distortion-free images within the dataset of distorted images as a form of adversarial augmentation. This strategy simulates the scenario of attack samples, challenging the model to maintain performance consistency. Given a distortion-free image, the ground truth pixel-wise flow map value equals the pixel coordinate and the sampling function outputs the distortion-free image itself. As shown in Table~\ref{table:aba_mix}, adding distortion-free images to the training or pre-training dataset slightly improves the performance of the model in all cases, especially for central fisheye images.

\begin{table}[!htbp] 
\centering
\caption{Comparison with the state-of-the-art methods on both deviated and central synthesized fisheye images}
\label{table:compare}
\begin{tabularx}{3in}{c|>{\centering\arraybackslash}X>{\centering\arraybackslash}X>{\centering\arraybackslash}X>{\centering\arraybackslash}X}
\toprule
\multirow{2}{*}{\textbf{Methods}} & \multicolumn{2}{c}{\textbf{Central  Image}} & \multicolumn{2}{c}{\textbf{Deviated  Image}} \\ 
\cmidrule{2-5}
  & \textbf{SSIM↑} & \textbf{PSNR↑} & \textbf{SSIM↑} & \textbf{PSNR↑} \\ 
\midrule
DR-GAN~\cite{liao-et-al:dr} & 0.7031 & 18.66 & 0.4875 & 14.66 \\
ModelFree~\cite{liao-et-al:model} & 0.6694 & 17.51 & 0.5226 & 14.98 \\
MLC~\cite{liao-et-al:multi} & 0.7345 & 19.33 & 0.5396 & 15.12 \\
PCN~\cite{yang-et-al:progressively} & 0.7687 & 20.53 & 0.5774 & 15.42 \\
SimFIR~\cite{feng-et-al:simfir} & 0.8612 & 22.47 & 0.5837 & 15.79 \\
DaFIR~\cite{liao-et-al:DaFIR} & 0.9053 & 26.30 & 0.5835 & 15.72 \\
Ours & \textbf{0.9162} & \textbf{26.38} & \textbf{0.9054} & \textbf{24.39} \\
\bottomrule
\end{tabularx}
\end{table}
\vspace{-0.1in}
\begin{table*}[!htbp] 
\centering
\caption{Ablation studies on the supplement of distortion-free images. N denotes distortion-free images.}
\label{table:aba_mix}
\vspace{-0.1in}
\begin{tabularx}{4.8in}{cc|>{\centering\arraybackslash}X>{\centering\arraybackslash}X|>{\centering\arraybackslash}X>{\centering\arraybackslash}X}
\toprule
\multicolumn{2}{c|}{\textbf{Dataset}} & \multicolumn{2}{c|}{\textbf{Central Fisheye Image}} & \multicolumn{2}{c}{\textbf{Deviated Fisheye Image}} \\
\cmidrule{1-6}
\textbf{Pre-training} & \textbf{Fine-tuning}  & \textbf{SSIM↑} & \textbf{PSNR↑} & \textbf{SSIM↑} & \textbf{PSNR↑} \\
\midrule
- & 100k-M & 0.8785 & 23.80 & 0.8343 & 21.70 \\
- & 80k-M+20k-N & 0.8974 & 24.59 & 0.8545 & 22.30 \\
400k-M & 100k-M & 0.9097 & 25.92 & 0.8944 & 23.89 \\
120k-C & 30k-C & 0.9053 & 26.05 & 0.5835 & 15.72 \\
280k-D & 70k-D & 0.7469 & 18.42 & 0.8905 & 23.45 \\
320k-M+80k-N & 80k-M+20k-N & \textbf{0.9162} & \textbf{26.38} & \textbf{0.9054} & \textbf{24.39} \\
\bottomrule
\end{tabularx}
\end{table*}

\begin{figure*}[t]
	\centering
	\includegraphics[width=2\columnwidth]{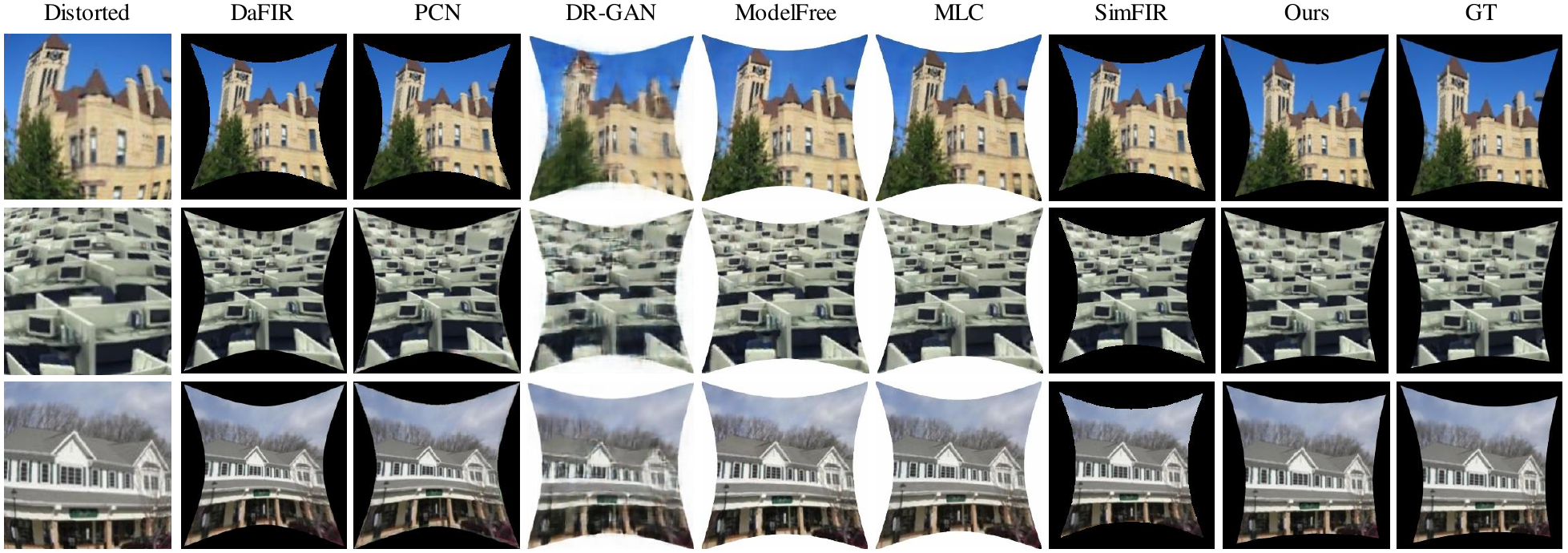}
	\caption{Qualitative comparison on the synthesized deviated fisheye images. From left to right, the sequence is as follows: the distorted image, DaFIR~\cite{liao-et-al:DaFIR}, PCN~\cite{yang-et-al:progressively}, DR-GAN~\cite{liao-et-al:dr}, ModelFree~\cite{liao-et-al:model}, MLC~\cite{liao-et-al:multi}, SimFIR~\cite{feng-et-al:simfir}, our method, and the ground truth.}
	\label{visual_compare5}
        \vspace{-0.1in}
\end{figure*}

\subsection{Comparison with State-of-the-art Methods}
Our comprehensive evaluation juxtaposes the rectification results of our methods with those of state-of-the-art counterparts on both synthesized and real fisheye images, including central and deviated ones, through quantitative and qualitative analyses.

\smallskip
\par
\textbf{Quantitative Comparison.}
To measure the performance of the methods, we analysis our proposed RoFIR and contemporary state-of-the-art rectification methods, including DR-GAN~\cite{liao-et-al:dr}, ModelFree~\cite{liao-et-al:model}, MLC~\cite{liao-et-al:multi}, PCN~\cite{yang-et-al:progressively}, SimFIR~\cite{feng-et-al:simfir}, and DaFIR~\cite{liao-et-al:DaFIR}.  For the latter three, SimFIR~\cite{feng-et-al:simfir}, DaFIR~\cite{liao-et-al:DaFIR}, and PCN~\cite{yang-et-al:progressively}, these models are trained using our fine-tuning dataset and evaluated on our test dataset. Conversely, for DR-GAN~\cite{liao-et-al:dr}, ModelFree~\cite{liao-et-al:model}, MLC~\cite{liao-et-al:multi}, these released models are directly assessed using our test dataset. As shown in Table~\ref{table:compare}, our method outperforms existing state-of-the-art methods both on central and deviated fisheye images.

\smallskip
\par
\begin{figure*}[t]
	\centering
	\includegraphics[width=2\columnwidth]{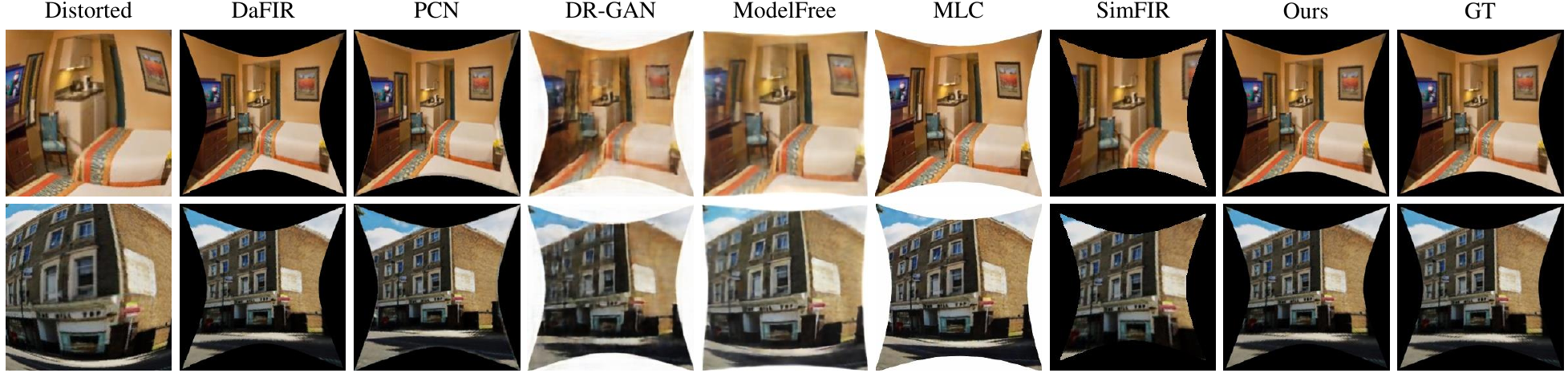}
          \vspace{-0.1in}
	\caption{Qualitative comparison on the synthesized central fisheye images. From left to right, the sequence is as follows: the distorted image, DaFIR~\cite{liao-et-al:DaFIR}, PCN~\cite{yang-et-al:progressively}, DR-GAN~\cite{liao-et-al:dr}, ModelFree~\cite{liao-et-al:model}, MLC~\cite{liao-et-al:multi}, SimFIR~\cite{feng-et-al:simfir}, our method, and the ground truth.}
	\label{visual_compare6}
         \vspace{-0.1in}
\end{figure*}

\textbf{Qualitative Comparison.}
For intuitiveness, this experiment qualitatively compares the performance of the RoFIR and state-of-the-art methods on both deviated and central fisheye images. We visualize the rectification results of deviated fisheye images using the aforementioned methods, as depicted in Figure~\ref{visual_compare5}. Existing methods exhibit a notable decline in the rectification performance on deviated fisheye images, with evident distorted structures remaining in the rectification results. Conversely, the RoFIR method demonstrates superior performance in the rectification of deviated fisheye images, in terms of both structural accuracy and detail quality. We also compare the rectification results on central fisheye images between the RoFIR method and existing methods. As shown in Figure~\ref{visual_compare6}, the RoFIR model also excels in performance on central fisheye image rectification. Notably, when the SSIM exceeds 0.85 and the PSNR exceeds 25, discerning differences in rectification results becomes difficult for humans. Both qualitative and quantitative experiment results reveal the effectiveness and robustness of the RoFIR in rectifying central and deviated fisheye images.

\smallskip
\par
\textbf{Comparison on Real-world Images.}
To evaluate the generalization ability, this experiment conducts a comparison of the rectification performance between the RoFIR method and existing state-of-the-art methods on real fisheye images. The test dataset encompasses real central fisheye images, as well as real deviated fisheye images. The real central fisheye images are captured by several fisheye lenses with different focal lengths and FoVs. The real deviated fisheye images are randomly cropped from the real central ones following the method described in Section~\ref{makedeviated}.

\begin{figure*}[t]
	\centering
	\includegraphics[width=2\columnwidth]{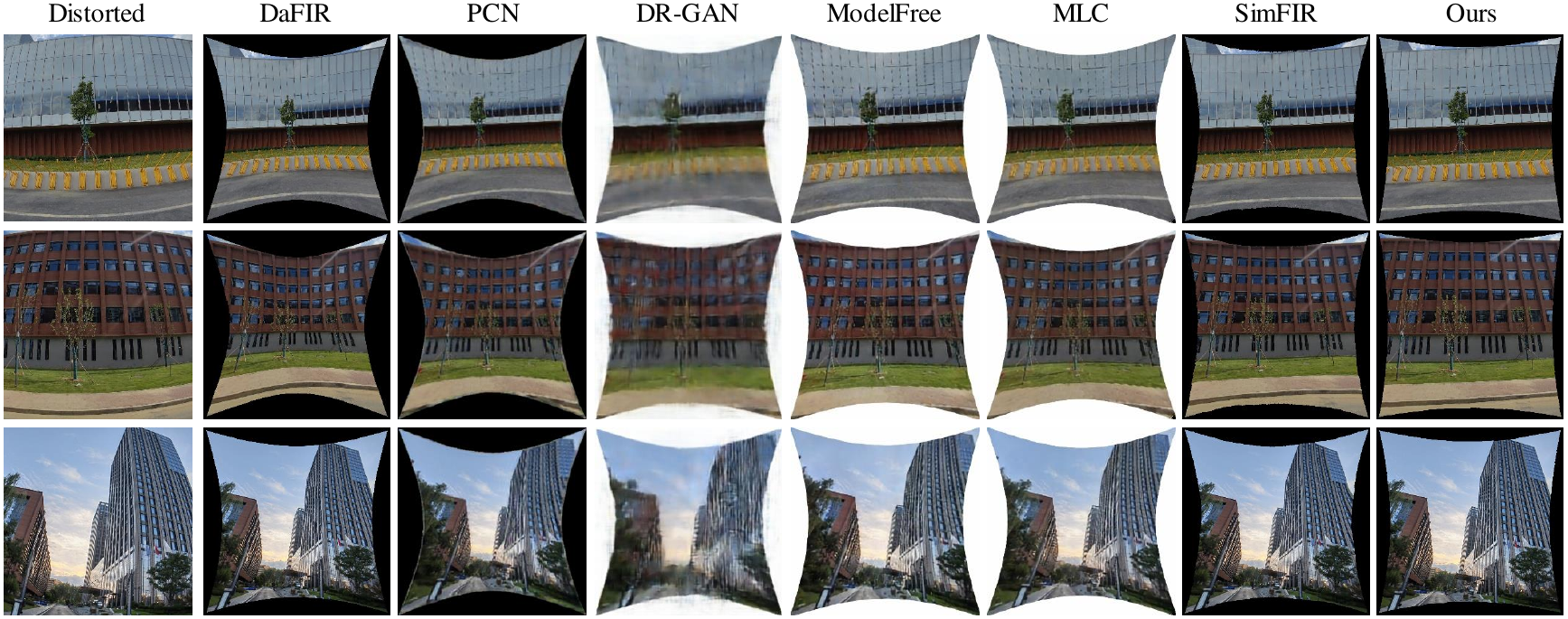}
	\caption{Qualitative comparison on the real deviated fisheye images. From left to right, the sequence is as follows: the distorted image, DaFIR~\cite{liao-et-al:DaFIR}, PCN~\cite{yang-et-al:progressively}, DR-GAN~\cite{liao-et-al:dr}, ModelFree~\cite{liao-et-al:model}, MLC~\cite{liao-et-al:multi}, SimFIR~\cite{feng-et-al:simfir}, and our method.}
	\label{visual_compare7}
        \vspace{-0.1in}
\end{figure*} 

\par The rectification results of real deviated fisheye images are illustrated in Figure~\ref{visual_compare7}. Methods including DaFIR~\cite{liao-et-al:DaFIR}, SimFIR~\cite{feng-et-al:simfir}, and PCN~\cite{yang-et-al:progressively} encounter a pronounced degradation in structural accuracy when rectifying real deviated fisheye images, despite their exemplary performance on real central fisheye images. The cause for this degradation is attributed to excessive reliance on the global distortion distribution patterns inherent in central fisheye images. Methods including DR-GAN~\cite{liao-et-al:dr}, ModelFree~\cite{liao-et-al:model}, and MLC~\cite{liao-et-al:multi} exhibit deficiencies in both structural accuracy and detail quality. The proposed RoFIR model retains high performance, evidencing strong generalization ability. This is attributed to the $\bm{V}$ maps that accurately capture local distortion features, allowing for rectification independent of global distortion patterns.

\par
\begin{figure*}[t]
	\centering
	\includegraphics[width=2\columnwidth]{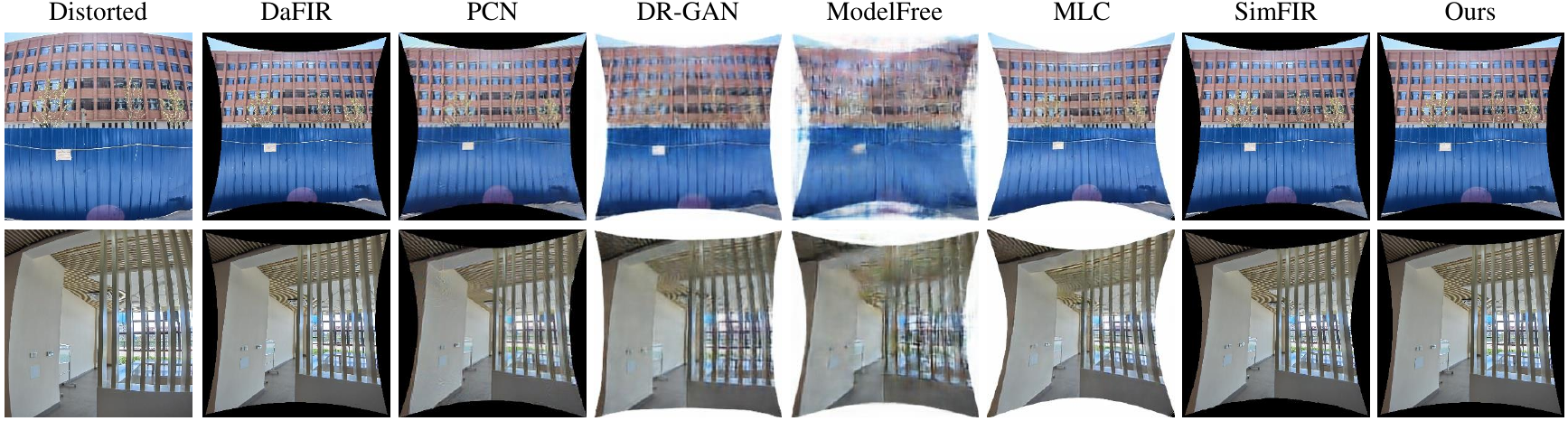}
	\caption{Qualitative comparison on the real central fisheye images. From left to right, the sequence is as follows: the distorted image, DaFIR~\cite{liao-et-al:DaFIR}, PCN~\cite{yang-et-al:progressively}, DR-GAN~\cite{liao-et-al:dr}, ModelFree~\cite{liao-et-al:model}, MLC~\cite{liao-et-al:multi}, SimFIR~\cite{feng-et-al:simfir}, and our method.}
	\label{visual_compare8}
        \vspace{-0.1in}
\end{figure*}

 The rectification results on real central fisheye images, illustrated in Figure~\ref{visual_compare8}, demonstrate that the RoFIR method is compatible with the rectification of central fisheye images and exhibits commendable generalization capabilities. The performance of RoFIR is on par with the SimFIR~\cite{feng-et-al:simfir} and DaFIR~\cite{liao-et-al:DaFIR} methods. The additional capability of the RoFIR method to rectify deviated fisheye images does not compromise its performance in rectifying central fisheye images. Experimental results indicate that the RoFIR model, trained on a mixed dataset, is effective for rectifying both central and deviated fisheye images, demonstrating functional versatility and robustness.

\section{Conclusion}

This paper raises the problem of rectifying deviated fisheye images with a randomly positioned optical center. This situation complicates rectification because it disrupts the consistent global distortion pattern, on which existing methods rely. To address this problem, we introduce a pixel-wise distortion vector map that measures the local distortion magnitude and direction. The proposed model undergoes two-stage training. In the pre-training stage, the model predicts distortion vector maps to perceive the local distortion patterns. During the fine-tuning stage, the model forecasts pixel-wise flow maps for rectification. Furthermore, our method is equally applicable to central fisheye images. A dataset mixing strategy, incorporating central, deviated, and distortion-free images, further enhances the rectification performance on both central and deviated fisheye images. Experimental results confirm that our method outperforms the competition in both function versatility and performance superiority.

\begin{acks}
To Robert, for the bagels and explaining CMYK and color spaces.
\end{acks}

\bibliographystyle{ACM-Reference-Format}
\bibliography{sample-base}










\end{document}